%% file: emnlp-ijcnlp-2019.tex
\documentclass[11pt,a4paper]{article}
\usepackage[hyperref]{emnlp-ijcnlp-2019}
\usepackage{times}
\usepackage{latexsym}

\usepackage{url}

\aclfinalcopy 

\aclfinalcopy

\title{Understanding BERT Performance in Propaganda Analysis}

\author{Yiqing Hua \\
  Cornell Tech \\
  {\tt yiqing@cs.cornell.edu} \\}

\date{}

\begin{document}
\maketitle
\begin{abstract}
\input{00-abstract.tex}
\end{abstract}

\input{01-introduction.tex}
\input{02-related.tex}
\input{03-methodology.tex}
\input{04-result.tex}

\input{06-acknowledgement.tex}

\bibliography{emnlp-ijcnlp-2019}
\bibliographystyle{acl_natbib}

\end{document}

%% file: 00-abstract.tex
In this paper, we describe our system used in the shared task for fine-grained propaganda analysis at sentence level. 
Despite the challenging nature of the task, our pretrained BERT model (team YMJA) fine tuned on the training dataset provided by the shared task scored 0.62 F1 on the test set and ranked third among 25 teams who participated in the contest.
We present a set of illustrative experiments to better understand the performance of our BERT model on this shared task.
Further, we explore beyond the given dataset for false-positive cases that likely to be produced by our system.
We show that despite the high performance on the given testset,
our system may have the tendency of classifying opinion pieces as propaganda and cannot distinguish quotations of propaganda speech from actual usage of propaganda techniques.

%% file: 01-introduction.tex
\section{Introduction}

The NLP4IF shared task for 2019 consists of 451 newspaper articles
from 48 news outlets that have been tagged for characteristics of 18 propaganda
techniques \cite{EMNLP19DaSanMartino}. 
The 18 propaganda techniques range from loaded language, name calling/labelling,
repetition, to logical fallacies such as oversimplification, red herring, etc.
Some of the techniques, by definition, require background knowledge to detect,
such as the identification of slogans, which would first require one to know of the slogans.

The shared task consists of two subtasks, sentence level classification~(SLC) and fragment level classification~(FLC).
In this paper, we focus our discussion on the sentence level classification. 
The subtask involves determining for each sentence, whether the text is `propaganda' or not as
a binary task.
The definition of being `propaganda' is that whether the utterance uses one of the 18 propaganda techniques listed in~\cite{EMNLP19DaSanMartino}.

In this paper, we describe our fine tuned BERT model used in the shared task. 
Our system (team YMJA) scored 0.62 F1 on the test set and ranked number third in the final 
competition for the SLC task. Further, we perform analyses in order to better understand the performance of our system.
Specifically, we would like to understand if the model was able to 
identify propaganda given appearances of the defined propaganda techniques,
or if it is exploiting obvious features that may lead to harmful false-positive examples.

Our results show that trained on the provided dataset from the shared task, our system may classify opinion pieces as propaganda and cannot distinguish quotation of propaganda speech from usage of propaganda techniques.
We advise that future applications of propaganda analysis algorithms trained with similar definition of propaganda should be used with caution. 
We hope the insights gained from our study can help towards the design of a more robust system in propaganda analysis. 

%% file: 02-related.tex
\section{Related Work}

Transformer based models \cite{vaswani2017attention}
such as BERT~\cite{DBLP:journals/corr/abs-1810-04805}
have swept the field of natural language process and has led to
reported improvements in virtually every task.
The capacity of these models to capture long term dependencies
and to represent context in ways useful for tagging is by
now well established with multiple papers suggesting
best practices \cite{DBLP:journals/corr/abs-1905-05583}.

Early works have applied machine learning techniques
directly to the problem of propaganda labelling at article level~\cite{rashkin-etal-2017-truth,volkova2018misleading,barron2019proppy},
with varying definitions of propaganda.
On the other hand, concerns have been raised regarding whether or not transformer
based models themselves \cite{solaiman2019release}
will lead to propaganda
generated by machines to deceive humans.
However, others have argued  \cite{DBLP:journals/corr/abs-1905-12616}
that strong fake news generators are an essential part of
detecting machine generated propaganda.

%% file: 03-methodology.tex
\section{Fine Tuned BERT models}
\label{sec:method}

In principle, this tagging problem in the NLP4IF  shared  task is similar to the well
studied problem of sentiment analysis for which there is ample
literature. Our own model draws heavily from a Kaggle 
competitor's
shared kernel \cite{yuval} built upon the popular
PyTorch Transformers libraries \cite{pytorch-transformers}.
While that application in \cite{yuval} is targeting toxicity in online
comments, a change of labels is sufficient to make the
same model apply to propaganda detection.
The code of our implementation can be found in the published colab file\footnote{\url{https://bit.ly/2kYmYwb}}.

We retrieved the uncased BERT-large model from github\footnote{\url{https://github.com/google-research/bert}}
and fine tuned the model on the training set.
We used 10-fold cross validation to create an ensemble of models.
The use of model ensemble techniques
\cite{DBLP:journals/corr/abs-1106-0257}
to limit over-training
and improve model performance on
held out test data is a well established. This is a common
feature of most Kaggle competitions, despite the fact that
the resulting models consume substantially more resources.

We trained each model for 1 epoch, with batch size of 32, learning rate of $10^{-5}$, decay of 0.01 and max sentence length of 129.
Given the imbalance of positive and negative labels, we up-weight positive samples with a factor of $5$ in the cross-entropy loss.

Our ensemble of models scored 0.62 F1 on the test set and ranked third among 25 teams, likely because the ensemble decreases the degree of overfiting.

%% file: 04-result.tex
\section{Discussion}
\label{sec:analysis}

\begin{table*}[thb]
    \centering
    \begin{tabular}{crrr}
         Method &  F1 & Precision & Recall\\
         \hline
         Perspective baseline & 0.57 & 0.54 & 0.60 \\
         Sentence length baseline & 0.47 & 0.44 & 0.51\\
         BERT ensemble & 0.66 & 0.63 & 0.69\\
    \end{tabular}
    \caption{Performance of baselines on development set.}
    \label{tab:res_on_dev}
\end{table*}

Recently, concerns have been raised 
\cite{DBLP:journals/corr/abs-1907-07355}
about the way that
transformer based models encode information about the world.
There exists a very real possibility that the answers to
questions about what statements are true or propaganda might
have been identifications of trivial statistical cues that
exist in the training data.
Therefore, 
in this section, we perform the following analysis in order to better understand whether our system ``understands'' the true nature of propaganda.

The two largest categories of propaganda techniques being used
are loaded language and name calling/labeling ($51\%$ according to ~\cite{EMNLP19DaSanMartino}).
Since these two techniques are also often used in the scenario of online harassment 
and trolling, we experiment with tools specialized in online harassment detection
in order to provide with a baseline.
To this end, we use Perspective API~\footnote{\url{https://github.com/conversationai/perspectiveapi/blob/master/api_reference.md}}.
Given an utterance and a defined attribute, the API returns a score between 0 to 1 as an estimate of the probability the utterance contains properties of the attribute.
The attributes are toxicity, severe toxicity, identity attack, insult,
profanity, threat, sexually explicit, flirtation,
inflammatory, obscene, likely to reject (by New York Times moderators) and unsubstantial.
The details of attributes' definitions are described at Perspective API website.
We aggregate these scores as sentence-level features and train a logistic regression on top of them to predict 
the likelihood of the sentence being propaganda.
As shown in Table~\ref{tab:res_on_dev},
the Perspective-API baseline achieves 0.57 F1, with 0.54 precision and 0.60 recall on 
development set, 
better accuracy than the provided baseline using sentence length.
Given that Perspective API was created for an unrelated task, the performance is surprisingly high and likely results from a high proportion of certain types of propaganda techniques in the dataset.

\begin{table*}[thb]
    \centering
    \begin{tabular}{lp{10cm}}
         \hline
         unigrams & devastating, cruel, vile, irrational, absurd, brutal, vicious, stupid, coward, awful, ignorant, unbelievable, doomed, idiot, terrifying, disgusting, horrible, hideous, horrific, pathetic\\ 
         \hline
         bigrams & shame less, totally insane, a horrible, utterly unacceptable, hysterical nonsense, the horrible, this horrific, absolutely disgusting, monumental stupidity, a pathetic, a disgusting, absolutely worthless, truly disgusting, utterly insane, this murderous, incredibly stupid, monstrous fraud, this lunatic, a disgrace, a hideous\\ 
         \hline
    \end{tabular}
    \caption{Top 20 unigrams and bigrams with highest likelihood of being propaganda.}
    \label{tab:unigram}
\end{table*}

In the second analysis we investigate the unigrams and bigrams being labeled with highest likelihood of being propaganda by our trained BERT model.
To this purpose, we fed all unigram and bigram combinations that appeared in the provided training set, development set and test set into our ensemble model
to infer their likelihood of being propaganda.
We list the top 20 unigrams and bigrams with highest probability of being propaganda determined by our system in Table~\ref{tab:unigram}.
Many of the shown terms indicate uncivil usage~(such as stupid, coward), or strong emotion~(such as terrifying, devastating, horrible).
In fact, the inclusion of such words in sentences would often lead to the sentence being classified as propaganda.
However, these combinations may as well be used in opinion pieces published in credible news sources.
Indeed, our system predicts certain titles of opinion pieces as propaganda with high likelihood.
For example, ``Devastating news for America's intelligence''\footnote{\url{https://www.washingtonpost.com/opinions/2019/08/02/devastating-news-americas-intelligence/}} published in Washington Post was scored with 0.85 probability of being propaganda by our system.
Given the definition of this shared task, it could be the intended behavior that opinion pieces being considered as propaganda.
Nevertheless, it is important to inform future users of this dataset and the resulting systems that opinion pieces are likely going to be classified as propaganda.

Another concern that this analysis raises is the limited capability of a system like this to distinguish quotations from actual usage of propaganda techniques.
News articles often have the need to quote original speech from political figures or other events, who might use techniques of propaganda.
Our analysis shows that the prediction of our system is not changed for a sentence when it is expressed as a quotation and that the mere presence of trigger words may lead to the classification of propaganda. 

In conclusion, the shared task on fine-grained propaganda analysis at NLP4IF workshop raises an important problem and with its dataset provides a key tool to analyze and evaluate progress.
However, as our analysis illustrated, there remains the challenge that the dataset appears unbalanced in that it focuses on loaded language and name calling/labelling. This makes it challenging for systems to capture signals of other more subtle or complex types of propaganda techniques.
For example, despite its high performance on the given test set, our BERT ensemble model trained on this dataset has high likelihood of failing in a real world scenario,
such as distinguishing quotations from actual propaganda.
We hope this study can help inform a more refined definition and a more diverse dataset towards propaganda analysis.

%% file: 06-acknowledgement.tex
\section{Acknowledgement}

We thank Jeffrey Sorensen and Andreas Veit for their helpful advices and proofreading.
We thank the anonymous reviewer for their suggestions.